\documentclass{article}
\pdfoutput=1
\PassOptionsToPackage{numbers, compress}{natbib}
\usepackage[preprint]{neurips_2020}

\usepackage[utf8]{inputenc} % allow utf-8 input
\usepackage[T1]{fontenc}    % use 8-bit T1 fonts
\usepackage{hyperref}       % hyperlinks
\usepackage{url}            % simple URL typesetting
\usepackage{booktabs}       % professional-quality tables
\usepackage{amsfonts}       % blackboard math symbols
\usepackage{nicefrac}       % compact symbols for 1/2, etc.
\usepackage{microtype}      % microtypography
\usepackage{graphicx}
\usepackage{accents}
\usepackage{caption}
\usepackage{subcaption}
\usepackage{verbatim}
\usepackage[ruled,vlined]{algorithm2e}
\usepackage{listings}

\title{Representaciones del aprendizaje reutilizando los gradientes de la retropropagación}
\author{%
  Roberto Reyes-Ochoa\\
  Departamento de Física\\
  Tecnológico de Monterrey\\
  Monterrey, NL  \\
  \texttt{a00344331@exatec.tec.mx} \\
  \And
  Servando López-Aguayo\\
  Departamento de Física\\
  Tecnológico de Monterrey\\
  Monterrey, NL  \\
  \texttt{servando@tec.mx}
}

\begin{document}

\maketitle

\begin{abstract}
  Este trabajo propone el algoritmo de gradientes de aprendizaje para encontrar significado en las entradas de una red neuronal. Además, se propone una manera de evaluarlas por orden de importancia y representar el proceso de aprendizaje a través de las etapas de entrenamiento. Los resultados obtenidos utilizan como referencia el conjunto de datos acerca de tumores malignos y benignos en Wisconsin. Esta referencia sirvió para detectar un patrón en las variables más importantes del modelo gracias, así como su evolución temporal.
\end{abstract}

\section{Introducción}

\begin{figure}[htbp]
    \centering
    \includegraphics[width=\linewidth]{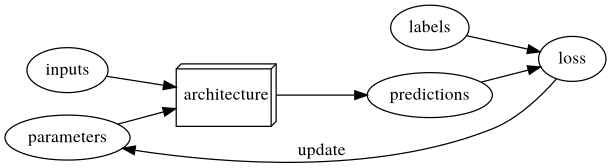}
    \caption{Modelo de redes neuronales profundas (DNN) en donde una arquitectura formada por varias capas de neuronas interconectadas reciben datos de entrenamiento (variable independiente) y parámetros. Posteriormente, utilizando estos parámetros se calcula una predicción para cada conjunto de datos. Estas predicciones son comparadas con el resultado objetivo (variable dependiente) y actualizan la función de pérdida. Finalmente, para minimizar esta función se actualizan los parámetros de la red iterativamente.}
    \label{fig:modelo}
\end{figure}

Las redes neuronales artificiales (ANNs, por sus siglas en inglés) son modelos computacionales que tratan de emular la forma en la que nuestro cerebro aprende a identificar patrones. A diferencia de un programa computacional típico, las ANNs no son explícitamente programadas a través de una serie de comandos, sino que utilizan un conjunto de parámetros cuyo valor es "aprendido" de acuerdo al tipo de tarea que estén desempeñando. Esta idea provino originalmente de Arthur Samuel, el padre del aprendizaje automático~\footnote{Cabe recalcar que existen otras técnicas de aprendizaje automático, tales como: redes bayesianas, árboles de decisión, SVMs, algoritmos de clustering, etc.}, quien habló de estas ideas en su ensayo \textit{"Artificial Intelligence: A Frontier of Automation"}. El siguiente diagrama (Fig. \ref{fig:modelo}) es una adaptación de las ideas de Arthur Samuel a la forma en la que se utilizan actualmente estos modelos.

Estos modelos computacionales han resultado sumamente eficaces para resolver problemas de naturaleza no lineal que tienen que ver con procesamiento de imágenes, lenguaje y sonidos. Por ejemplo, las redes neuronales convolucionales (CNNs), son arquitecturas particularmente buenas para clasificar e identificar patrones en imágenes, a tal grado que han reemplazado al ser humano en tareas como la lectura de caracteres escritos a mano. Por otro lado, las redes neuronales recurrentes (RNNs) han conseguido resultados impresionantes al momento de generar textos originales y traducir idiomas. Uno de los ejemplos actuales más impresionantes es el modelo GPT-3 de NVIDA, el cual utiliza cerca de un billón de parámetros para producir cadenas de texto originales a partir de un tema de conversación. Otros modelos más sofisticados, como las GAN (\textit{Generative Adversarial Networks}) han aprovechado la capacidad discriminativa de las CNNs para construir redes neuronales que generen rostros artificiales realistas.

\subsection{¿Cómo funciona una ANN?}

La unidad estructural de una red neuronal es la neurona. El concepto de neurona proviene de la naturaleza biológica de nuestro cerebro. En la naturaleza, una neurona recibe una serie de señales de entrada y puede producir o no una señal de salida en función de la suma de intensidades de las señales entrantes . Este comportamiento es emulado por las neuronas artificiales y se conoce como función de activación. Se busca que esta función de activación cumpla con ciertas características como ser no-lineal, continuamente diferenciable, monotónica y de rango fijo. Por ello, tradicionalmente la función de activación por excelencia había sido la función sigmoidal (Eq. \ref{eq:activacion}).

\begin{equation}\label{eq:activacion}
    f(z) = \frac{1}{1+e^{-z}}
\end{equation}

Sin embargo, en la práctica la función \textit{ReLU} (Eq. \ref{eq:relu}) arroja mejores resultados cuando se utiliza en las capas intermedias de las redes neuronales profundas. La simplicidad de esta función podría llevarnos a dudar de su utilidad; sin embargo, es una función que cumple con la no-linealiadad y es mucho más fácil de procesar, de manera que, en la práctica, los modelos que utilizan esta función convergen más rápido a la solución.

\begin{equation}\label{eq:relu}
    f(z) = \max{(0, z)}
\end{equation}

\begin{figure}
\centering
\begin{subfigure}{.5\textwidth}
  \centering
  \includegraphics[width=\linewidth]{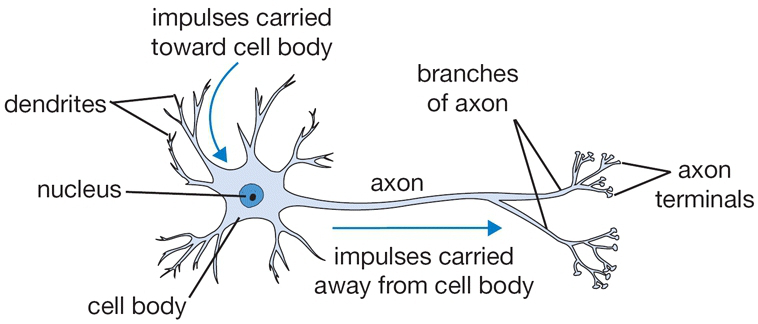}
  \caption{Modelo biológico}
  \label{fig:neurona_bio}
\end{subfigure}%
\begin{subfigure}{.5\textwidth}
  \centering
  \includegraphics[width=\linewidth]{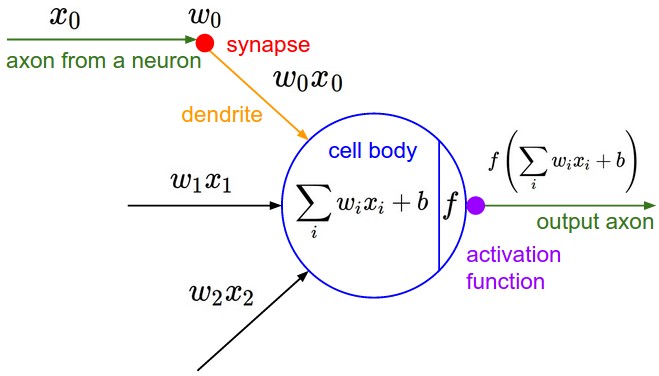}
  \caption{Modelo matemático}
  \label{fig:neurona_mate}
\end{subfigure}
\caption{Analogía entre el modelo biológico y el matemático de una neurona. En (b) se observa cómo la función de activación $f$ se activa cuando la suma de impulsos $\sum_iw_ix_i$ supera el umbral de activación $-b$. Normalmente, al término $\sum_iw_ix_i + b$ se le asigna la variable $z$ para simplificar la notación, tal que la activación se pueda escribir como $f(z)$.}
\label{fig:neuronas}
\end{figure}

Así pues, la activación de una neurona marca la pauta para la activación subsecuente de otra neurona que recibe como entrada este valor (Fig. \ref{fig:neuronas}). Este proceso se repite de manera iterativa en una red neuronal, desde la capa de entrada en donde las activaciones son ni más ni menos que las entradas de la red, pasando por las capas intermedias (también conocidas como capas ocultas) en donde las neuronas reciben activaciones de las neuronas de capas anteriores, hasta finalmente llegar a la capa de salida, donde las activaciones representan las predicciones del modelo.

\begin{figure}
    \centering
    \includegraphics[width=0.8\linewidth]{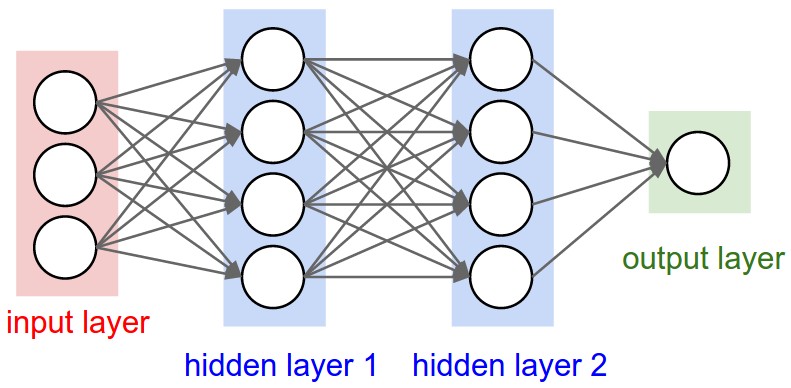}
    \caption{Ejemplo de la arquitectura de una ANN totalmente conexa con dos capas ocultas. La representación matemática de la neurona que está en la capa de salida es: $\hat{y}=f(\sum_kw_kf_k(\sum_jw_jf_j(\sum_iw_ix_i + b_j) + b_k) + b)$, donde $i$ corresponde al número de neuronas en la capa de entrada, $j$ al de la primera capa oculta y $k$ al de la segunda capa oculta.}
    \label{fig:ann}
\end{figure}

Las interconexiones neuronales están acompañadas de dos parámetros: pesos y umbrales (que en inglés se conocen como \textit{bias}). Cada peso está asignado a un par de neuronas diferente y su función es priorizar algunas activaciones de la capa anterior sobre las otras. Por otro lado, existe un umbral por cada neurona de las capas intermedias y de la capa de salida, su función es establecer el punto de activación de la neurona. La Fig. \ref{fig:ann} muestra lo que se conoce como una ANN totalmente conectada, es decir, para cada neurona de la capa siguiente existe una conexión con todas las neuronas de la capa anterior. Como se menciona en la descripción de la figura, cada representación gráfica puede acompañarse de una representación matemática para cada neurona $n$ de la capa de salida $L$ (Ec. \ref{prediccion}). Donde $x_k$ representa el conjunto de entradas para un ejemplo de entrenamiento, $w_{jk}^{(l)}$ las combinaciones de pesos para la capa $l$ (en donde $j$ es el índice que recorre las neuronas de capa actual y $k$ el de las de la capa anterior), $b_j^{(l)}$ son los umbrales de cada neurona $j$ de la capa $l$ y $f_k^{(l)}(...)$ el conjunto de activaciones de cada neurona $k$ de la capa $l$. Esta ecuación se puede construir aplicando $f$ de manera recursiva desde la última capa, pasando por las capas $L-1$, $L-2$, ..., hasta llegar a la capa de entrada.

\begin{equation}\label{prediccion}
    \hat{y}_n(x_k,w_{jk}^{(l)},b_j^{(l)}) = f_n^{(L)} \left(\sum_kw_{jk}^{(L-1)}f_k^{(L-1)} \left(\sum_kw_{jk}^{(L-2)}\dots + b_j^{(L-1)}\right) + b_j^{(L)}\right)
\end{equation}

El objetivo de una ANN es modificar estos dos tipos de parámetros para que dada una entrada $x_k$ se produzca una respuesta $\hat{y}_n$ en la capa de salida que sea lo más parecida a la salida esperada. La forma en la que se consigue este objetivo es utilizando ejemplos de entrenamiento para optimizar una función de costo. Los ejemplos de entrenamiento son una serie de observaciones $(x_k^{(m)}, y_n^{(m)})$ en donde $y_n^{(m)}$ representa la salida observada para el vector de entradas $x_n^{(m)}$, mientras que $m$ indica el ejemplo de entrenamiento~\footnote{Una de las grandes ventajas de las redes neuronales es que mejoran considerablemente conforme incrementa el número de ejemplos de entrenamiento disponibles}. Por otro lado, la función de costo es una función heurística que compara el resultado de la red $\hat{y}_n^{(m)}$ con la $y_n^{(m)}$ observada, como lo indica el modelo de la Fig. \ref{fig:modelo}. Esto sirve como una heurística para saber qué tan bien se está desempeñando la red. Un ejemplo es la función MSE (\textit{Mean Squared Error}) que se muestra en la Eq. \ref{loss}, donde $C$ es el costo promedio de evaluar $m$ ejemplos de entrenamiento, $n$ son los componentes de la salida que se suman y $m$ los ejemplos de entrenamiento que se promedian.

\begin{equation}\label{loss}
    C(\hat{y}_n^{(m)}) = \frac{1}{2m}\sum_m\sum_n(\hat{y}_n^{(m)}-y_n^{(m)})^2
\end{equation}

% hablar de feedforward
En el contexto de redes neuronales, al proceso de buscar los parámetros que optimizan la función de costo se le conoce como entrenamiento. Existen dos etapas fundamentales para este proceso: propagación y retro-propagación. La primera de ellas involucra tomar la entrada $x_k^{(m)}$ de un ejemplo de entrenamiento $(x_k^{(m)}, y_n^{(m)})$ y propagarlo a través de toda la red, de capa en capa, de manera que las entradas activan las neuronas de la segunda capa y éstas activan las de la tercera, así hasta llegar a la capa de salida en donde se produce el vector de salidas $\hat{y}_n^{(m)}$. La función de costo recibe entonces estas salidas y las compara con la salida observada $y_n^{(m)}$ para saber qué tan lejos está la red de alcanzar la convergencia. 

% backprop
Por otro lado, la etapa de retropropagación o \textit{backprop}, consiste en calcular los $\delta{w}$ y $\delta{b}$ a partir de la función de costo. La forma intuitiva de pensar en este proceso es imaginar una red de una sola capa. Iniciando en la capa de salida se procede en reversa: primero obtenemos el cámbio que se busca en la función de costo, pasando por el cambio en la función de activación, el cambio necesario de los umbrales y terminando en el cambio en los pesos. Concluimos el proceso actualizando los parámetros de acuerdo a estos gradientes calculados. Si existen más capas, simplemente se van reutilizando los gradientes hasta llegar a los parámetros de la capa de entrada. Este proceso se repite hasta agotar los ejemplos de entrenamiento (la Fig. \ref{alg:retroprop} representa la aplicación de este algoritmo a un ejemplo de entrenamiento con $k$ entradas y $j$ salidas). Una vez que se acaban los ejemplos de entrenamiento concluye la primera etapa de entrenamiento. El número de etapas de entrenamiento, así como el número de ejemplos de entrenamiento utilizados por cada paso del algoritmo de retropropagación (\textit{mini-batches}) son hiperparámetros de la red que alteran el proceso global de minimización llamado SGD (\textit{Stochastic Gradient Descent}).

\begin{algorithm}[htbp]
\SetAlgoLined
\KwData{Ejemplo de entrenamiento $(x_k, y_j)$, conjunto de pesos $w^{(l)}_{jk}$ y umbrales $b^{(l)}_j$}
\KwResult{Gradientes $\delta{w^{(l)}_{jk}}$ y $\delta{b^{(l)}_{j}}$ que actualizan los parámetros $w^{(l)}_{jk}$ y $b^{(l)}_j$ para el ejemplo $(x_k, y_j)$}
\BlankLine
$\hat{y}_i, z^{(l)}_i, a^{(l)}_i \longleftarrow FeedForward(x_i, w^{(l)}_{jk}, b^{(l)}_j)$\;
$\delta{C}_j \longleftarrow (\hat{y}_j-y_j) f'_j(z_j)$\;
$\delta{w^{(L)}_{jk}} \longleftarrow \delta{C}_{j}a^{(L-1)}_{k}$\;
$\delta{b^{(L)}_{j}} \longleftarrow \delta{C}_{j}$\;
$l \longleftarrow L-1$\;
\While{l >= 2}{
  $\delta{C}^{(l)}_{k} \longleftarrow f'_k(z_k^{(l)}) \sum_j\delta{C}^{(l+1)}_{j} \delta{w}^{(l+1)}_{jk}$\;
  $\delta{w^{(l)}_{jk}} \longleftarrow \delta{C}^{(l)}_{j}a^{(l-1)}_{k}$\;
  $\delta{b^{(l)}_{j}} \longleftarrow \delta{C}^{(l)}_{j}$\;
  $l \longleftarrow l - 1$\;
}
\caption{Obtener gradientes de retropropagación que actualizan los parámetros}
\label{alg:retroprop}
\end{algorithm}

Los hiperparámetros son parámetros intrínsecos del modelo que no se optimizan con el entrenamiento, pero que determinan el potencial inicial de la red para resolver un problema de optimización determinado. Algunos ejemplos son: número de etapas de entrenamiento, tamaño de \textit{mini-batch}, inicialización de los parámetros, preprocesamiento de las entradas, entre otros. De hecho, Un tema que se dejó de lado intencionalmente hasta este momento y que forma parte del conjunto de hiperparámetros es la arquitectura de la red neuronal. Para empezar, la arquitectura de una red neuronal artificial está restringida en la capa de entrada y en la capa de salida. Sin embargo, la capa intermedia ofrece un sin fin de variantes.

%\subsection{¿Por qué funciona?}

% limitaciones de la analogía biológica

% interpretación: hablar de cómo las redes neuronales son aproximadores de funciones. 

% cómo los pesos y el bias modifican la función de activación para clasificar correctamente los ejemplos de entrenamiento

% hablar sobre mínimos locales, learning rate, dropout y la normalización de las entradas

\subsection{Propuesta de investigación}

Generalmente, los gradientes de retropropagación sólo se utilizan para optimizar los parámetros de las redes neuronales; sin embargo, en este trabajo propongo utilizarlos para destacar las propiedades de la entrada a partir del error en la salida. Esto con el propósito de alcanzar un mayor entendimiento cualitativo del proceso de aprendizaje de este tipo de modelos. ¿Con qué objetivos específicos?
\begin{enumerate}
    \item Simplificación de la entrada: identificar la importancia de cada variable en el modelo, considerando que esto es de gran utilidad en la simplificación de modelos complejos de muchas variables.
    \item Representación del aprendizaje: entender qué es lo que está aprendiendo la red neuronal conforme pasan las etapas de entrenamiento. Es decir, cómo evoluciona el entendimiento de las entradas conforme se actualizan los parámetros o se modifican los hiperparámetros.
\end{enumerate}

\section{El algoritmo de gradientes de aprendizaje}

% lo que importa es identificar a qué le presta atención la red

% impresión: la respuesta de la red a un conjunto de entradas, la modificación sobre las entradas que mejoraría el desempeño de la red

El paso clave del algoritmo de gradientes de aprendizaje es sumamente sencillo: almacenar los gradientes retropropagados hasta la capa de entrada, como si las entradas fueran también parámetros de la red. Como se observa en la Fig. \ref{alg:grad}, el paso final almacena los gradientes, los cuales forman un vector de la misma dimensión que la entrada. Esto permite que cada entrada pueda ser mapeada con un gradiente de aprendizaje.

\begin{algorithm}[htbp]
\SetAlgoLined
\KwData{Ejemplo de entrenamiento $(x_k, y_j)$, conjunto de pesos $w^{(l)}_{jk}$ y umbrales $b^{(l)}_j$}
\KwResult{Representaciones del aprendizaje de la red neuronal sobre el conjunto de entradas $x_i$}
\BlankLine
$\hat{y}_i, z^{(l)}_i, a^{(l)}_i \longleftarrow FeedForward(x_i, w^{(l)}_{jk}, b^{(l)}_j)$\;
$\delta{C}_j \longleftarrow (\hat{y}_j-y_j) f'_j(z_j)$\;
$\delta{w^{(L)}_{jk}} \longleftarrow \delta{C}_{j}a^{(L-1)}_{k}$\;
$\delta{b^{(L)}_{j}} \longleftarrow \delta{C}_{j}$\;
$l \longleftarrow L-1$\;
\While{l >= 2}{
  $\delta{C}^{(l)}_{k} \longleftarrow f'_k(z_k^{(l)}) \sum_j\delta{C}^{(l+1)}_{j} \delta{w}^{(l+1)}_{jk}$\;
  $\delta{w^{(l)}_{jk}} \longleftarrow \delta{C}^{(l)}_{j}a^{(l-1)}_{k}$\;
  $\delta{b^{(l)}_{j}} \longleftarrow \delta{C}^{(l)}_{j}$\;
  $l \longleftarrow l - 1$\;
}
$gradiente_k \longleftarrow norm_k(abs_k(\sum_j\delta{C}^{(2)}_{j} \delta{w}^{(2)}_{jk}))$
\caption{Obtener gradientes de aprendizaje a partir de retropropagación}
\label{alg:grad}
\end{algorithm}

Los gradientes del aprendizaje se relacionan con aquellas entradas a las cuales la red le presta mucho mayor atención. Un par de detalles importantes: los gradientes están expresados en magnitud y son normalizados, esto con el objetivo de clasificar cualitativamente la importancia de cada variable y responder la primera pregunta de investigación. De manera rigurosa, estos gradientes corresponden con la modificación sobre las entradas que mejoraría el desempeño de la red sobre un ejemplo de entrenamiento (aunque también puede ser un \textit{mini-batch} de ejemplos) dados todos los parámetros actuales.

Identificar la lista de factores más influyentes al concluir todas las etapas de entrenamiento es sólo una de las posibles aplicaciones de los gradientes de aprendizaje. Con ellos también es posible identificar la evolución del aprendizaje. El procedimiento es simplemente comparar los gradientes correspondientes a cada variable conforme transcurren las etapas de entrenamiento. Este análisis puede ser incluso más granular considerando los gradientes correspondientes a cada \textit{mini-batch} o, incluso, a cada ejemplo de entrenamiento. Sin embargo, al agregar granularidad se observa una pérdida de generalización como se describe en \cite{alqaraawi2020}.

% una segunda interpretación está en relacionar este conjunto de valores con la interpretación de las entradas que más valora la red

\subsection{Generalización a otras arquitecturas}

El algoritmo de gradientes de aprendizaje se puede generalizar a otras arquitecturas como CNN. Algunos autores han reportado el uso de apas de saliencia~\cite{alqaraawi2020, simonyan2014, mundhenk2020} y redes deconvolucionales que generalizan la propagación de gradientes a través de las capas convolucionales y de pooling \cite{zeiler2013, springenberg2015}. 
% mencionar la extensión para CNN y RNN
% existe versión para RL?

\section{Resultados}

Primero, estaremos usando el conjunto de datos de resultados de cancer en Wisconsin importado de la librería \texttt{sklearn}. Este dataset es lo suficientemente sencillo para experimentar con diferentes formas de representación de aprendizaje usando retropropagación sin necesidad de introducir conceptos de redes convolcionales.

La hipótesis es que una red totalmente conexa de entre 1 y 3 capas ocultas podrá alcanzar una precisión aceptable para esta tarea de clasificación binaria. Este modelo, una vez entrenado, permitirá examinar las representaciones del aprendizaje.

Posteriormente, los modelos de representación encontrados podrán generalizarse y obervarse de manera más clara en redes convolucionales usando imágenes. Esto sería un proceso análogo a los mapas de saliencia.

\subsection{Casos de cancer en Wisconsin}

Esta es la descripción detallada del conjunto de datos proporcionada por \texttt{sklearn}.

\begin{lstlisting}
Breast cancer wisconsin (diagnostic) dataset
--------------------------------------------

**Data Set Characteristics:**

    :Number of Instances: 569

    :Number of Attributes: 30 numeric, predictive attributes and
    the class

    :Attribute Information:
        - radius (mean of distances from center to points on 
        the perimeter)
        - texture (standard deviation of gray-scale values)
        - perimeter
        - area
        - smoothness (local variation in radius lengths)
        - compactness (perimeter^2 / area - 1.0)
        - concavity (severity of concave portions of the contour)
        - concave points (number of concave portions of the contour)
        - symmetry 
        - fractal dimension ("coastline approximation" - 1)

        The mean, standard error, and "worst" or largest (mean of 
        the three largest values) of these features were computed 
        for each image, resulting in 30 features.  For instance, 
        field 3 is Mean Radius, field 13 is Radius SE, field 23 is
        Worst Radius.

        - class:
                - WDBC-Malignant
                - WDBC-Benign
\end{lstlisting}

\begin{figure}[htbp]
    \centering
    \includegraphics[width=\linewidth]{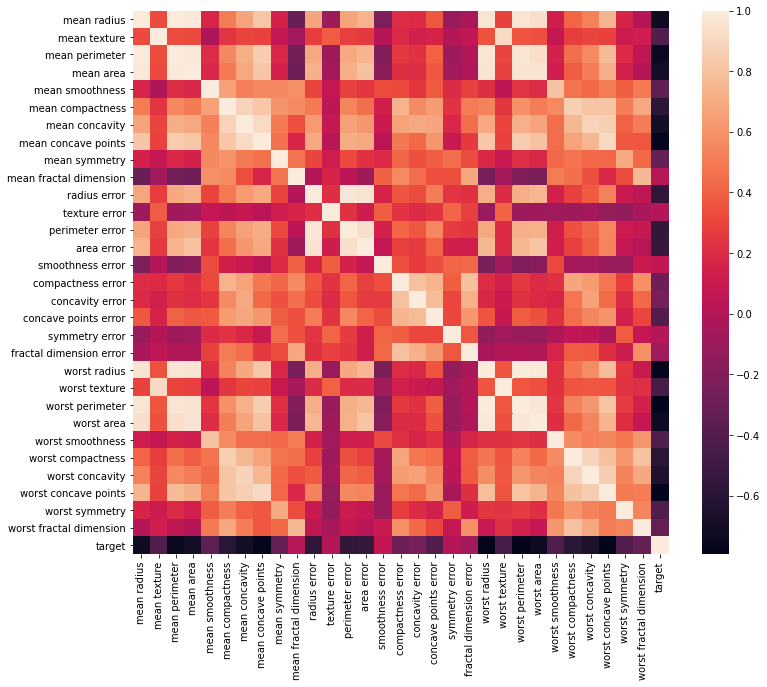}
    \caption{Variables del conjunto de datos de cancer en Wisconsin y sus coeficientes de correlación.}
    \label{fig:corr}
\end{figure}

Como punto de referencia, consideramos que 0 representa un tumor maligno y 1 significa tumor benigno. La Fig. \ref{fig:corr} muestra de manera cuantitativa cómo se relacionan las variables de este conjunto de datos.

Es evidente que existen grupos de variables que se correlacionan muy fuertemente. Esto se debe principalmente a que algunas de las variables son estadísticos como la media y el error que están evaluados sobre la misma variable, como, por ejemplo, el radio del tumor. Acto seguido se procede a construir el dataset de entrenamiento y de prueba en el formato que facilite el entrenamiento de la red usando la librería \texttt{pytorch}.

Es importante mencionar que las arquitecturas simples de redes neuronales del tipo \texttt{feedforward} requieren que las variables de entrada sean estandarizadas, esto para evitar que las variables de mayor escala adquieran una importancia desproporcionada. El procedimiento que realiza la estandarización es simplemente restar el promedio de cada variable y dividirlo entre su desviación estándar.

Posteriormente, se procede a construir el modelo de red neuronal, pensando en la inicialización de los parámetros (pesos y umbrales) de cada capa, las funciones de propagación para cada capa y su correspondiente mapeo no lineal. A partir de una optimización de hiperparámetros, se optó por una arquitectura completamente conexa con dos capas ocultas: $m\times 3 \times t$, donde $m=30$ es el número de entradas y $t=1$ es el número de variables de respuesta. Es decir, una arquitectura tipo perceptrón con una capa oculta de 3 neuronas.

\begin{figure}
    \centering
    \includegraphics[width=\linewidth]{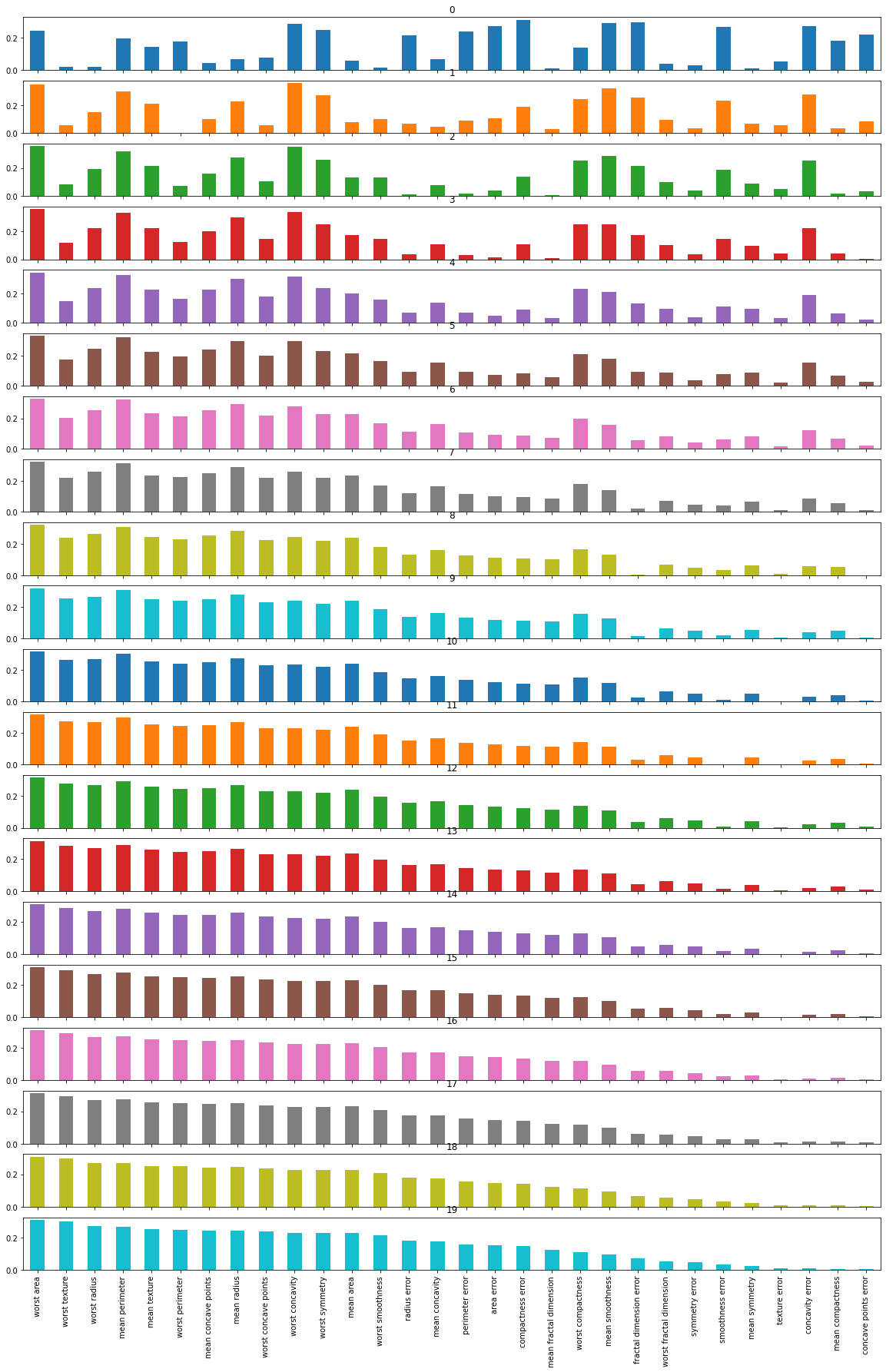}
    \caption{Modelo con ajuste de 100\% sobre el conjunto de prueba después de 40 etapas de entrenamiento. Las 5 variables más importantes después de este periodo de entrenamiento fueron: (1) \texttt{worst area}, (2) \texttt{worst texture}, (3) \texttt{worst radius}, (4) \texttt{mean perimeter} y (5) \texttt{mean texture}.}
    \label{fig:evol}
\end{figure}

Finalmente, se implementó el algoritmo \ref{alg:grad} embebido en el proceso de optimización SGD. Después de pasar el modelo por 40 etapas de entrenamiento se obtuvo una precisión de 100\% sobre el conjunto de prueba y posteriormente se realizó un experimento utilizando los gradientes obtenidos. Estos resultados se observan en la Fig. \ref{fig:evol}, donde cada gráfica corresponde a una etapa de entrenamiento impar y las barras corresponden a un componente de la distribución generada por los gradientes, la cual asigna un porcentaje de relevancia a cada variable de entrada. 

% presentar un caso, intentar primero con handwriten digits

% observar la distribución de las entradas

% entrenar la red

% regresar la lista de impresiones

% observar la distribución de la impresión correspondiente a las entradas

% variar el número de neuronas por capa y de capas ocultas

\section{Discusión}

Previamente se ha hablado acerca de las representaciones del aprendizaje, principalmente entorno a redes convolucionales \cite{zeiler2013, springenberg2015, simonyan2014}. Estos avances trajeron herramientas tecnológicas que utilizaron estas intuiciones para interpretar la importancia de las entradas de redes neuronales en general (como por ejemplo \url{captum.ai}). Este trabajo propone en cambio utilizar los gradientes de retropropagación con el objetivo principal de representar el aprendizaje codificado en los parámetros e hiperparámetros de la red.

Los resultados obtenidos mostraron una convergencia en los gradientes de retropropagación observados conforme pasaban las etapas de entrenamiento. Considerando las limitaciones de las redes neuronales artificiales, esto implica que bajo cierto óptimo local la red considera que existen variables más importantes que otras al momento de calcular la función de error. La Fig. \ref{fig:evol} logró el objetivo de representar la evolución de la representación del aprendizaje.

Como trabajo futuro se abordará a detalle la evolución del aprendizaje en redes convolucionales, pues existe la posibilidad de identificar cómo se aprenden diferentes secciones de la imagen conforme transcurren los periodos de entrenamiento.

\section{Código}

Código en python para la construcción del modelo, entrenamiento y resultados sobre el dataset de cáncer en Wisconsin: \url{https://colab.research.google.com/drive/1BtG3Ih6k3cbBWvv1U84n_QUCkZ0D3lBH?usp=sharing}

\bibliographystyle{unsrtnat}

\end{document}